%% file: icml2026.tex
\documentclass{article}

\usepackage{microtype}
\usepackage{graphicx}
\usepackage{subcaption}
\usepackage{booktabs} 

\usepackage{hyperref}
\usepackage{diagbox}

\usepackage[preprint]{icml2026}

\usepackage{amsmath}
\usepackage{amsmath}
\usepackage{amssymb}
\usepackage{mathtools}
\usepackage{amsthm}
\usepackage{tabularray}
\usepackage{makecell}
\usepackage{multirow}
\usepackage[capitalize,noabbrev]{cleveref}
\usepackage{comment}
\theoremstyle{plain}

\theoremstyle{definition}

\theoremstyle{remark}

\usepackage{colortbl}
\usepackage{tikz}
\usetikzlibrary{tikzmark, backgrounds, fit}
\usepackage{xcolor}

\definecolor{stage2color}{RGB}{240, 245, 255}

\usepackage[textsize=tiny]{todonotes}

\icmltitlerunning{Submission and Formatting Instructions for ICML 2026}

\newcommand{\sys}{C-SAM}

\begin{document}

\twocolumn[
  \icmltitle{Towards Compact and Robust DNNs via \\ Compression-aware Sharpness Minimization}



  \icmlsetsymbol{equal}{*}

  \begin{icmlauthorlist}
    \icmlauthor{Jialuo He}{yyy}
    \icmlauthor{Huangxun Chen}{yyy}

  \end{icmlauthorlist}

  \icmlaffiliation{yyy}{The Hong Kong University of Science and Technology (Guangzhou)}

  \icmlcorrespondingauthor{Huangxun Chen}{huangxunchen@hkust-gz.edu.cn}

  \icmlkeywords{Machine Learning, ICML}

  \vskip 0.3in
]



\printAffiliationsAndNotice{}  

\input{section/0-abs}
\input{section/1-intro}

\input{section/4-related}

\input{section/2-design}
\input{section/3-eval-set}
\input{section/3-eval-res}

\input{section/5-conclude}

\newpage
\section*{Impact Statement}
The proposed \sys{} framework addresses a critical gap in deploying reliable AI on edge devices under stringent memory and storage constraints. 
By enabling DNNs to remain both compact and certifiably robust against real-world semantic variations, this research has strong potential to make advanced AI more accessible and resilient in privacy-sensitive edge applications, including facial user authentication, in-home activity monitoring, daily smart logging for personal assistants, etc, where natural input variations are commonplace. 
Overall, \sys{} reconciles model compactness with robustness, ensuring that robustness attained during training can be effectively preserved to the pruned models used in deployment. 
\bibliography{icml2026}
\bibliographystyle{icml2026}

\newpage
\appendix
\onecolumn
\input{section/6-appendix}


\end{document}

%% file: section/0-abs.tex
\begin{abstract}

Sharpness-Aware Minimization (SAM) has recently emerged as an effective technique for improving DNN robustness to input variations. However, its interplay with the compactness requirements of on-device DNN deployments remains less explored.
Simply pruning a SAM-trained model can undermine robustness, since flatness in the continuous parameter space does not necessarily translate to robustness under the discrete structural changes induced by pruning. 
Conversely, applying SAM after pruning may be fundamentally constrained by architectural limitations imposed by an early, robustness-agnostic pruning pattern.
To address this gap, we propose \textbf{C}ompression-aware \textbf{S}h\textbf{A}rpness \textbf{M}inimization (\sys{}), a framework that shifts sharpness-aware learning from parameter perturbations to \emph{mask perturbations}. 
By explicitly perturbing pruning masks during training, \sys{} promotes a flatter loss landscape with respect to model structure, enabling the discovery of pruning patterns that simultaneously optimize model compactness and robustness to input variations. Extensive experiments on CelebA-HQ, Flowers-102, and CIFAR-10-C across ResNet-18, GoogLeNet, and MobileNet-V2 show that \sys{} consistently achieves higher certified robustness than strong baselines, with improvements of up to 42\%, while maintaining task accuracy comparable to the corresponding unpruned models.

\end{abstract}

%% file: section/1-intro.tex
\section{Introduction}

DNN robustness has attracted increasing research attention. 
A model is considered robust if it maintains accurate and consistent outputs when its inputs are subject to natural variations encountered in real-world settings, a property closely tied to its generalization capability. 
Recently, many studies have leveraged Sharpness-Aware Minimization (SAM)~\cite{sharpness2, sharpness3, sharpness4, sharpness5, sharpness6, sharpness1} to enhance model robustness. 
The underlying philosophy is that a smoother, flatter loss landscape correlates with stronger generalization~\cite{hessian,hessian2,losslandscape}. Accordingly, SAM explicitly promotes flat loss landscape during training by {injecting stochastic perturbations into model parameters} and minimizing the task loss. 

Despite SAM's effectiveness in improving the robustness of a given DNN architecture, applying it to on-device DNNs for edge scenarios introduces new challenges. 
The memory-bound nature of edge devices often necessitates compressing a full DNN into a lightweight variant, yet typical edge applications, such as facial user authentication and in-home activity monitoring, still demand strong model robustness in real-world deployments. 

A seemingly intuitive approach is to prune a SAM-trained model. However, SAM is formulated over a dense and continuous parameter space, whereas pruning induces discrete structural changes. As a result, a model that lies in a flat region with respect to weight-magnitude perturbations may still be highly sensitive to structural connectivity changes, and its robustness is therefore not guaranteed to be preserved after pruning. 
Recent work, S$^2$-SAM~\cite{s2sam}, can be used to fine-tune the pruned model with SAM to enforce robustness. In addition, AdaSAP \cite{adasap} adopts a progressive strategy that alternates between pruning to an intermediate sparsity level and applying SAM, iterating this process until the target pruning ratio is reached. However, in these pipelines, the pruning stage remains largely agnostic to robustness objectives, and the achievable robustness may be fundamentally bounded by pruning-induced structural constraints. 
We also notice that some earlier efforts, e.g., HYDRA, Flying Bird \cite{hydra, fbp}, reconciled robustness with compactness primarily through adversarial training under pixel-level $\ell_p$-norm perturbation assumptions \citep{fgsm,pgd}. 
Consequently, compared with SAM, the resulting models are less robust to realistic input variations that arise along the underlying data manifold rather than in pixel space, such as changes in illumination, weather conditions, or facial expressions~\citep{gcert, genprove}. 
Therefore, a natural research question arises: \emph{Is it feasible to identify an effective pruning pattern that simultaneously optimizes model compactness and robustness to input variations?} 

We provide a positive answer to this question by proposing \textbf{C}ompression-aware \textbf{S}h\textbf{A}rpness \textbf{M}inimization (\sys{}) framework. 
We adopt the general principle of SAM, however, instead of perturbing model parameters, we explicitly introduce mask variables to encode the compactness objective and perturb these masks during training.  
Shifting the perturbation from model parameters to pruning masks allows us to explicitly enforce \emph{structural flatness}, i.e. flatter loss landscape in pruning mask space, enabling the discovery of subnetworks that remains robust to both input variations and pruning-induced structural changes. 

To materialize our idea, the key step is to define an effective objective over the mask variables that drives the learning of a pruning pattern achieving both the target pruning ratio and strong robustness.
Specifically, \sys{} features a composite objective that jointly accounts for structural stability, input semantic robustness and deployment consistency, as illustrated in \cref{smile}. 
We first introduce a stability loss $\mathcal{L}_{\mathrm{stab}}$ to minimize the prediction variance induced by stochastic mask perturbations. 
By enforcing consistency across similar pruning patterns, $\mathcal{L}_{\mathrm{stab}}$ effectively smooths the loss landscape over model structure and ensures that similar subnetworks yield consistent predictions (\cref{fig:losslandscape}(b)). 
In addition, we explicitly account for the discrepancy between the mask variables (soft masks with continuous values) and the deployed masks (hard masks with binary values), and introduce a loss term $\mathcal{L}_{\mathrm{consis}}$ to enforce their consistency and prevent performance degradation during inference-time deployment. 
Last but not least, we derive a mathematical sufficiency condition that enables the design of a rigorous margin-aware ratio loss, $\mathcal{L}_{\mathrm{ratio}}$, to promote robustness against targeted input variations effectively generated by state-of-the-art diffusion models~\cite{diffae, Diffusionclip}. 
Incorporating these losses jointly steers training toward a structurally flat region of the loss landscape, ultimately yielding a compact yet robust DNN. 
Our key contributions are summarized as follows:

\noindent$\bullet$ We introduce the {structural flatness} notion for jointly optimizing DNN pruning and robustness, and operationalize it via mask-space perturbations and stability-loss regularization. 

\noindent$\bullet$ We propose \sys{}, a {compression-aware sharpness-minimization training} framework that optimizes pruning masks via a carefully designed composite objective, effectively bridging the gap between discrete model pruning and continuous robustness optimization. 

\noindent$\bullet$ 
Extensive experiments across multiple benchmarks (CelebA-HQ \cite{celeba}, Flowers-102 \cite{flowers}, and CIFAR-10-C \cite{cifar10,cifar10-c}) and DNN architectures (ResNet-18 \cite{resnet}, MobileNet-V2 \cite{mobilenetv2}, and GoogLeNet \cite{googlenet}) demonstrate that compared with state-of-the-art baselines, \sys{} identifies pruning patterns that meet the compactness target while achieving substantially higher certified robustness.  
Notably, on CelebA-HQ, \sys{} improves certified robustness by up to 42\% and even surpasses unpruned models by 8\%–27\% at a 50\% compression ratio, while maintaining comparable task accuracy (\cref{tab:googlemobile}).

%% file: section/4-related.tex
\section{Related Work}\label{sec:relatedwork}

\textbf{Model Compression.} 
Model compression aims to reduce a model’s memory footprint. Pruning is one of the effective approaches among others and has been extensively studied, including \emph{unstructured} weight sparsification \citep{unstr1,unstr2,unstr3,unstr4,unstr5,unstr6,unstr7} and \emph{structured} channel/filter removal \citep{depgraph,hesso,otov1,otov2,str5,str6,str4,str2,str3,llmprune,llmprune2}.

\textbf{Model Robustness.} 
Model robustness is crucial for real-world deployment, as models must maintain accurate and consistent outputs even when inputs undergo natural variations in the operating environment. 
In our work, we adopt the more rigorous notion of certified robustness~\cite{cccert,sok_cr} as our optimization objective. 

\textbf{Compact and Robust DNNs.}
On-device DNNs must be compact to satisfy stringent memory/storage constraints, while remaining robust to input variations. 
Some early efforts integrate adversarial training with model compression~\cite{hydra,fbp,harp,atmc,ip1-3,ip1-1,ip1-2,ip1-4,ip1-5,ip1-6}. HYDRA \cite{hydra} employs a learnable soft mask optimized to minimize adversarial loss.
Flying Bird \cite{fbp} jointly optimizes network weights and connectivity topologies. 
However, they assume $\ell_p$-bounded, pixel-level perturbations, which limits their effectiveness against natural semantic variations.
Besides, SAM has emerged as a more effective approach for enhancing general robustness \cite{sharpness1,sharpness2,sharpness3,sharpness4,ssam,sharpness5,sharpness6}. 
Its key idea is to explicitly encourage convergence to flat minima in the loss landscape over model parameters. 
Some recent efforts have attempted to integrate SAM with pruning. 
\cite{s2sam} fine-tunes the pruned model with SAM, mitigates the chaotic loss landscapes inherent in sparse networks, and derives theoretical bounds on the generalization error. 
\cite{adasap} adopts a progressive strategy that alternates between pruning to an intermediate sparsity level and applying SAM, demonstrating robustness against out-of-distribution shifts, such as pixel-level Gaussian noise or fog corruptions \cite{cifar10-c}. 
However, pruning and robustness are optimized in consecutive, separated stages and remain largely agnostic to each other. Thus, the attainable robustness may be fundamentally bounded by pruning-induced structural constraints.
In contrast, \sys{} enables compression-aware robustness optimization by driving the model toward flat minima in a loss landscape over structural parameters, i.e., pruning masks.

%% file: section/2-design.tex
\section{Design}
\label{sec:method}
In this section, we first provide the necessary preliminaries, followed by a detailed description of the \sys{} framework.

\subsection{Preliminary}
\label{preliminary}
Given a classification task represented by the dataset $\mathcal{S}=\{(x_i,y_i)\}_{i=1}^m$, $x_i\in\mathbb{R}^n$ and $y_i\in\{1,\dots,K\}$. We employ a DNN-based classifier $f$, parameterized by $\theta$, such that $f: \mathbb{R}^n \to [0,1]^K$ maps each input to a probability distribution over $K$ classes. The final prediction is determined by the class with the highest probability, defined as $\hat{y}(\cdot) \triangleq \arg\max f(\cdot)$.

Following Chernoff-Cramer Certification (CC-Cert) framework \cite{cccert}, for an input $x$ with true class $c$ and transform $T:\mathbb{R}^n\to\mathbb{R}^n$, we obtain a space of transformed inputs $\mathbb{S}_T(x)$,  under $T$. The model $f$ is said to be probabilistically robust at $x$ with confidence $1-\varepsilon$ if the consistency of the prediction $\hat{y}$ over $\mathbb{S}_T(x)$ satisfies
\begin{equation}\label{1-epsilon}
\mathbb{P}_{x_T\sim\mathbb{S}_T}\!\Big(\hat{y}(x_T)=c\Big)\ \ge\ 1-\varepsilon.
\end{equation}

To operationalize this, we quantify the prediction discrepancy between $x$ and its transformed counterpart $x_T\sim\mathbb{S}_T(x)$ as
\begin{equation}\label{eq:z}
Z(x;T)\ \triangleq\ \|\mathbf{p}(x)-\mathbf{p}(x_T)\|_\infty
\end{equation}
where $\mathbf{p}(x)=f(x)$ denotes the class-probability vector, and we copy entries from $\mathbf{p}(x)$ and sort them in descending order, $p_1>p_2>\cdots>p_K$ to measure the margin between the top-2 predictions of input $x$ as $d(x)\triangleq\frac{p_1-p_2}{2}$. 

Thus, assuming the model correctly classifies input $x$, a sufficient condition for satisfying probabilistic robustness defined in \cref{1-epsilon} is $Z(x;T) < d(x)$, and we provide the complete proof in~\cref{proof:lableinvariance}. 
To summarize, increasing the certified passing probability is a surrogate for optimizing probabilistic robustness 
\begin{equation}
\label{eq:cpp}
\max~\mathbb{P}\big(Z(x;T)< d(x)\big)
\end{equation}
which can be achieved by minimizing the prediction discrepancy $Z(x;T)$ or by maximizing the margin $d(x)$. 

Following \citet{genprove,gcert,rs_semantic2}, let $G: z\in\mathcal{Z} \to x\in\mathcal{M}$ be a generator that approximates the natural image manifold $\mathcal{M}$ by mapping a lower-dimensional latent vector $z \in \mathcal{Z}$ to the image space. Formally, given an input $x = G(z)$ and a target semantic attribute associated with a unit direction vector $\mathbf{v} \in \mathbb{R}^d$, the mutated image $x_{T}$ is generated as:$$x_{T} = T(x, \delta, \mathbf{v}) = G(z + \delta \cdot \mathbf{v})$$where $\delta \in [0, 1]$ represents the magnitude of the semantic shift, e.g., the intensity of smiling. The set of all semantically mutated images is defined as $\mathbb{S}_T(x) = \{G(z+\delta \cdot \mathbf{v}) \mid \delta \in[0,1]\}$. An example of a transformation space is provided in \cref{fig:ccert_grid}.

For model compression, \sys{} is applicable to both structured and unstructured pruning. Let $\mathcal{P}$ denote the set of minimal prunable units (e.g., an individual weight for unstructured pruning or a channel for structured pruning), with $N = |\mathcal{P}|$ representing the total number of such units.
We introduce a learnable soft mask $C \in [0,1]^N$ as a continuous surrogate. To apply this mask to the weights, we define a broadcasting operator $\mathcal{T}: \mathbb{R}^N \to \mathbb{R}^{|\theta|}$, resulting in the compressed model $f_{C}(\cdot) = f(\cdot; \mathcal{T}(C) \odot \theta)$. 
Specifically, for unstructured pruning, $N = |\theta|$ and $\mathcal{T}$ reduces to the identity mapping. In contrast, for structured pruning, $N = C_{\text{out}}$ and $\mathcal{T}$ acts as a broadcasting operator, where $C_{\text{out}}$ denotes the number of output channels in CNNs \cite{str_1, str_2}.

Predefined pruning ratio ($\text{pr}$) can be written as: $\text{pr} = 1 - {\|\theta_{\hat{m}}\|_0}/{|\theta|}$,
where $\theta_{\hat{m}}=\mathcal{T}(\hat{m})\odot\theta$ denotes the active parameters of the deployed model. The binary mask $\hat{m} \in \{0, 1\}^N$ is derived by projecting $C$ onto the target density $k$, where $k = 1 - \text{pr}$. 
Specifically, we prune the model in a layer-wise way with a layer-wise threshold $\tau_k^i$ such that the top-$k$ proportion of elements in the $i$-th layer of $C$ are preserved:
\begin{equation}\label{eq:hardmask}
    \hat{m}^i = \mathbf{1}(C^i \ge \tau_k^i), \quad \text{s.t.} \quad \|\theta_{\hat{m}}\|_0 \le k\cdot|\theta|
\end{equation}
where $\mathbf{1}(\cdot)$ is the indicator function.

\subsection{Our Approach}
In this section, we first introduce the initialization method for the soft mask $C$, followed by a detailed discussion of the composite loss components: (1) $\mathcal{L}_\text{stab}$, (2) $\mathcal{L}_\text{ratio}$, and (3) $\mathcal{L}_\text{consis}$.

\noindent\textbf{Percentile-Scaled Initialization.}
To facilitate an effective mask search, we adopt a percentile-scaled initialization strategy. Instead of naive random initialization, the soft mask values are set proportional to the magnitudes of the pre-trained weights, thereby ensuring that critical connections are prioritized at the beginning of training. Specifically, for the $i$-th layer, the soft mask $C^i$ is initialized as follows:
\begin{equation}\label{eq:softmask}
C^i = \text{clip}\left( \frac{|\theta_i|}{Q(|\theta_i|, \tau)}, 0, 1 \right),
\end{equation}
where $Q(|\theta_i|, \tau)$ denotes the $\tau$-th percentile of the weight magnitudes in the $i$-th layer. Consequently, weights within the top $\tau\%$ are assigned a value of $1$, while the remaining entries are smoothly scaled within the range $[0, 1)$.

After initialization, the soft mask $C$ is treated as a stochastic variable to facilitate exploration of the loss landscape. Specifically, we generate individual mask instances by injecting uniform noise, followed by a clipping operation to maintain the values within the feasible range:
\begin{equation}\label{eq:mask}
    C_\xi = \text{clip}(C + \xi, 0, 1), \quad \xi \sim \mathcal{U}(-\mu, \mu),
\end{equation}
where $\mu$ denotes the hyperparameter controlling the magnitude of the injected noise.

\noindent\textbf{Flat Minima over Mask and Semantic Consistency.}
As shown in \cref{eq:cpp}, minimizing the prediction discrepancy $Z$ directly enhances model robustness. For a specific mask instance $C$, the upper bound of $Z$ can be derived via triangle decomposition as follows:
\begin{align}\label{eq:triangle}
Z_{C}(x;T) & \le \underbrace{\sqrt{\big\| \mathbf{p}_{C}(x) - \bar{\mathbf{p}}(x) \big\|_2^2}}_{\mathrm{(A)}}
+ \underbrace{\big\| \bar{\mathbf{p}}(x) - \bar{\mathbf{p}}(x_T) \big\|_\infty}_{\mathrm{(B)}} \nonumber\\
&\quad + \underbrace{\sqrt{\big\| \bar{\mathbf{p}}(x_T) - \mathbf{p}_{C}(x_T) \big\|_2^2}}_{\mathrm{(C)}},
\end{align}
where $\bar{\mathbf{p}}(x)=\mathbb{E}_{\xi}[\mathbf{p}_{C_\xi}(x)]$ denotes the \emph{ensemble mean} over stochastic mask instances. The complete derivation is provided in \cref{appendix:proof_Z}. The upper bound formulated in \cref{eq:triangle} consists of three synergistic components: Terms (A) and (C) quantify the prediction variance induced by stochastic compression masks, whereas Term (B) characterizes the model's intrinsic sensitivity to semantic transformations.

We decompose the optimization into two synergistic objectives: (1) stability optimization, which explicitly minimizes Terms (A) and (C) via $\mathcal{L}_{\text{stab}}$ to steer the model toward a flatter loss region; and (2) margin-aware semantic optimization, which targets Term (B) via $\mathcal{L}_{\text{ratio}}$ to enforce semantic consistency.

\begin{figure}[t]
    \centering
    \includegraphics[width=0.8\columnwidth]{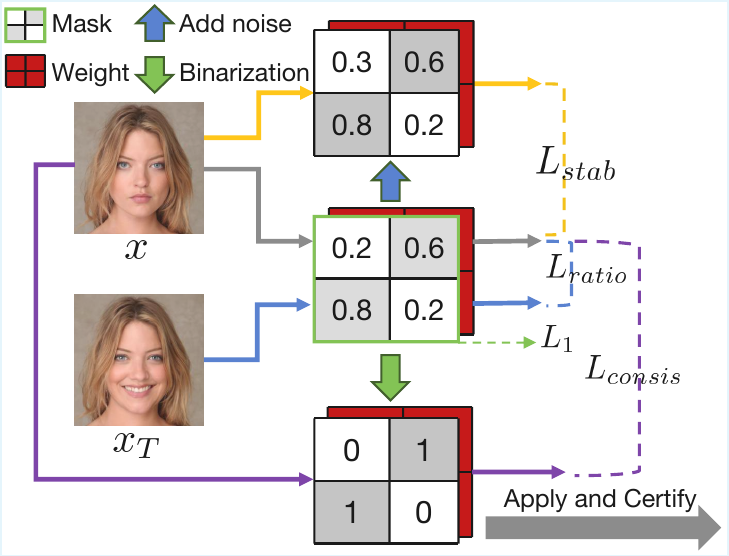}\vspace{-0.cm}
    \caption{\footnotesize An illustration of the \sys{} framework, where pruning masks are optimized for stability under mask perturbations ($\mathcal{L}_{\text{stab}}$), robustness against semantic input variations ($\mathcal{L}_{\text{ratio}}$), soft-hard mask consistency ($\mathcal{L}_{\text{consis}}$), and $L_1$ regularization.}\vspace{-0.1in} 
    \label{smile}
\end{figure}

\begin{algorithm}[htb!]
    \caption{Training Procedure of \sys{}}
    \label{alg:car}
    \begin{algorithmic}[1]
    \REQUIRE DNN $\theta$, dataset $\mathcal{S}$, transformation $T(\cdot, \delta)$, \\ pruning ratio $\text{pr}$, broadcasting operator $\mathcal{T}$, \\
    safety threshold $\eta$, small constant $\epsilon$, \\ coefficients $\lambda_\text{\{stab,consis,ratio,1\}}$, learning rate $\alpha$. 
    
    \STATE Construct augmented dataset $\mathcal{S}' = \mathcal{S} \cup \{(T(x,\delta), y) \mid (x,y) \in \mathcal{S}\}$, paired dataset $\mathcal{B}_p = \{(x, T(x,\delta))\}$ derived from $\mathcal{S}'$
    
    \item[] {\textcolor{gray}{\# Stage 1: Pre-training}}
    \STATE Train full-precision model on augmented dataset: $\theta_{\text{pre}} \leftarrow \arg\min_{\theta} \mathbb{E}_{(x,y)\sim\mathcal{S}'}[\mathcal{L}_{\mathrm{CE}}(\theta, x, y)]$

    \item[] {\textcolor{gray}{\# Stage 2: Robust Mask Search}}
    \STATE Initialize soft mask $C$ for each layer via \cref{eq:softmask} 
    \WHILE{not converged}
        \STATE Sample a batch: $(x, x_{{T}}) \leftarrow \mathcal{B}_p$
        \item[] \quad \textcolor{gray}{\textit{// 1. Stability Loss}}
        \STATE Generate stochastic masks $C_m, C_n, C_s$ by injecting noise into $C$ via \cref{eq:mask} 
        \STATE Probabilistic outputs: $\mathbf{p}_{C_m} \leftarrow f(x; \mathcal{T}(C_m)\odot\theta_{\text{pre}} )$, $\mathbf{p}_{C_n} \leftarrow f(x; \mathcal{T}(C_n)\odot\theta_{\text{pre}})$
        \STATE $\mathcal{L}_{\text{stab}} \leftarrow \| \mathbf{p}_{C_m} - \mathbf{p}_{C_n} \|_2^2$
        \item[] \quad \textcolor{gray}{\textit{// 2. Soft-Hard Consistency}}
        \STATE Compute binarization threshold $\tau_k^i$ to satisfy pr
        \STATE Binarize masks: $\hat{m}^i \leftarrow \mathbf{1}(C^i \ge \tau_k^i)$
        \STATE STE: $\hat{m}\leftarrow\big(\hat{m}- C\big).detach()+C $ 
        \STATE Hard output: $\mathbf{p}_{\hat{m}} \leftarrow f(x; \mathcal{T}(\hat{m})\odot\theta_{\text{pre}})$
        \STATE $\mathcal{L}_{\text{consis}} \leftarrow \text{KL}(\mathbf{p}_{C_m}\| \mathbf{p}_{\hat{m}} )$ 
        \item[] \quad \textcolor{gray}{\textit{// 3. Ratio Loss}}
        \STATE Semantic output: $\mathbf{p}_{C_s} \leftarrow f(x_T; \mathcal{T}(C_s)\odot\theta_{\text{pre}})$
        \STATE Calculate discrepancy $Z \leftarrow \|\mathbf{p}_{C_m}(x) - \mathbf{p}_{C_s}(x_T)\|_\infty$ 
        \STATE Calculate margin $d\leftarrow \frac{p_1-p_2}{2}$, where $p_1, p_2$ are top 2 entries of $\mathbf{p}_{C_m}$
        \STATE $\mathcal{L}_{\text{ratio}} \leftarrow \operatorname{Softplus}( \frac{Z}{d + \epsilon} - \eta )$
        \item[] \quad \textcolor{gray}{\textit{// 4. Update Mask}}
        \STATE $\mathcal{L} \leftarrow \lambda_{\text{stab}}\mathcal{L}_{\text{stab}} + \lambda_{\text{consis}}\mathcal{L}_{\text{consis}} + \lambda_{\text{ratio}}\mathcal{L}_{\text{ratio}} + \lambda_{1}\mathcal{L}_1$
        \STATE Update mask $C \leftarrow C - \alpha \nabla_{C} \mathcal{L}$
    \ENDWHILE

    \item[] {\textcolor{gray}{\# Stage 3: Post-training}}
    \STATE Compute binarization threshold $\tau_k^i$ to satisfy pr
    \STATE Binarize masks: $\hat{m}^i \leftarrow \mathbf{1}(C^i \ge \tau_k^i)$ to meet $\text{pr}$
    \STATE Fine-tune model parameters $\theta$ from $\theta_\text{pre}$: \\  $\theta_{f} \leftarrow \arg\min_{\theta} \mathbb{E}_{(x,y)\sim\mathcal{S}'}[\mathcal{L}_{\mathrm{CE}}(\mathcal{T}(\hat{m})  \odot \theta , x, y)]$
    \ENSURE Final compressed model $\mathcal{T}(\hat{m})  \odot \theta_f$
    \end{algorithmic}
\end{algorithm}

\textit{(1) Stability Loss}. Considering two independent stochastic mask instances $C_m$ and $C_n$ generated via the stochastic noise injection process (\cref{eq:mask}), the stability loss is formulated as follows: 
\begin{align}
\mathcal L_{\mathrm{stab}}
& \triangleq\ \mathbb E_x\,\mathbb E_{C_m,C_n}\big[\|\mathbf p_{C_m}(x)-\mathbf p_{C_n}(x)\|_2^2\big] \nonumber\\
& = 2\,\mathbb E_x\,\mathbb E_{\xi}\big[\|\mathbf p_{C_\xi}(x)-\bar{\mathbf p}(x)\|_2^2\big]
\label{eq:stabloss}
\end{align} 
The full proof is provided in \cref{variance identity}. Minimizing $\mathcal{L}_{\text{stab}}$ effectively reduces the prediction variance induced by stochastic compression masks, i.e., Terms (A) and (C), which helps tighten the theoretical upper bound on the prediction discrepancy $Z$. In essence, $\mathcal{L}_{\mathrm{stab}}$ serves as a regularization term that penalizes inconsistent predictions across different subnetwork topologies, thereby steering the optimization toward a structurally flatter region of the loss landscape.

\textit{(2) Ratio Loss}. Relying solely on $\mathcal{L}_{\text{stab}}$ would leave the semantic gap in Term (B) unconstrained. To address this, we propose ratio loss $\mathcal{L}_{\text{ratio}}$, which directly penalizes large semantic prediction discrepancies $Z(x; T)$ relative to the classification margin $d(x)$. Building on~\cref{eq:cpp}, we define the robustness ratio as: 
\begin{equation}
    r(x;T) \triangleq \frac{Z(x;T)}{d(x)+\epsilon} ,
\end{equation}
where $\epsilon>0$ is a small constant introduced to prevent division by zero.

To satisfy the certification condition, it is essential to penalize the robustness ratio whenever $r(x;T)>1$. Accordingly, we introduce a safety threshold $\eta \in (0,1]$ and employ the $\operatorname{softplus}$ function to implement a smooth penalty when the ratio exceeds this threshold:
\begin{equation}\label{eq:ratioloss}
\mathcal{L}_{\mathrm{ratio}}
\ \triangleq\
\mathbb{E}_{x,C_m,C_s}\Big[\operatorname{softplus}\big(r(x;T)-\eta\big)\Big].
\end{equation}
In practice, to prevent overfitting to a specific subnetwork, the discrepancy $Z(x;T)$ within $r(x;T)$ is computed using two independently sampled stochastic masks $C_m$ and $C_s$.

\noindent\textbf{From Continuous to Binary.} 
While continuous soft masks facilitate differentiable optimization, practical deployment necessitates discrete binary masks. To mitigate the performance degradation typically induced by binarization, we enforce behavioral alignment between the soft and hard masks during the training process.

\textit{(3) Consistency Loss.} 
Specifically, we minimize the divergence between the predictions generated by a stochastic soft mask $C_m$ and those from its binarized counterpart $\hat{m}$. This alignment is operationalized through the Kullback–Leibler (KL) divergence loss:
\begin{equation}\label{eq:consisloss}
\mathcal{L}_{\mathrm{consis}}
\triangleq\ \mathbb{E}_{x,C_m} \Big[\mathrm{KL}\big(\mathbf{p}_{C_m}(x)\|\mathbf{p}_{\hat{m}}(x)\big)\Big].
\end{equation}
To optimize this objective, we employ the Straight-Through Estimator (STE) \cite{ste_bengio} to bypass the non-differentiable binarization step, enabling gradients to directly update the underlying soft mask. This scheme ensures that the robustness properties acquired in the continuous optimization space are effectively preserved within the discrete model used for inference.

\noindent\textbf{Putting it All Together.} 
In addition to the objectives defined above, we incorporate an $\ell_1$ regularization term on the soft mask $C$ to induce sparsity, denoted as $\mathcal{L}_1\triangleq \|C\|_1$. The overall objective function is formulated as: 
\begin{equation}\label{eq:all}
\mathcal{L} =
\lambda_{\mathrm{stab}}\,\mathcal{L}_{\mathrm{stab}}
+\lambda_{\mathrm{ratio}}\,\mathcal{L}_{\mathrm{ratio}}
+ \lambda_{\mathrm{consis}}\mathcal{L}_{\mathrm{consis}}
+\lambda_{1}\mathcal{L}_1.
\end{equation}
The training pipeline of \sys, summarized in \cref{alg:car}, proceeds in three stages. First, we pre-train the full-precision model using standard cross-entropy loss $\mathcal{L}_\text{CE}$ to establish discriminative semantic features. Next, we perform a robust mask search. In this stage, we freeze the backbone parameters and optimize only the soft masks using $\mathcal{L}$. Finally, we binarize the resulting masks and fine-tune the compressed network to recover accuracy.

%% file: section/3-eval-set.tex
\section{Experiments Setup}
\label{sec:overalldesign}

\noindent\textbf{Datasets.} 
We evaluate our approach on CelebA-HQ \cite{celeba} and Flowers-102 \cite{flowers} using latent-space semantic mutations, and extend the evaluation to CIFAR-10-C \cite{cifar10,cifar10-c} to assess scalability against common corruptions. We augment these three datasets with two levels of data scale: L1 and L2. Specifically, for CelebA-HQ and Flowers-102, L1 and L2 consist of 1,000 and 4,000 samples, respectively. For CIFAR-10-C, the corresponding L1 and L2 levels are scaled to 10,000 and 20,000 samples. 
It is worth noting that all pruning baselines are retrained from scratch on the same augmented dataset $\mathcal{S}’$ (i.e., with semantic variations) as ours to ensure a fair comparison. 
Detailed descriptions of the augmentation process are provided in \cref{appendix:data}.

\noindent\textbf{Evaluation Models.} We validate our approach using ResNet-18 \citep{resnet}, GoogLeNet \citep{googlenet}, and MobileNet-V2 \citep{mobilenetv2}. We select these architectures to represent lightweight model families commonly deployed on edge devices, rather than large-scale, over-parameterized networks. 

\noindent\textbf{Comparison Baselines.}\label{Comparison with baselines} 
To isolate the impact of compression, we first evaluate two unpruned models: {Vanilla Train}, trained with standard cross-entropy loss, and {AdvTrain}, which incorporates adversarial loss \cite{pgd}. Subsequently, we categorize the pruning baselines into unstructured and structured approaches, with comprehensive descriptions and configurations provided in \cref{appendix:baseline}.

\textit{Unstructured Pruning.} 
We select representative methods from three distinct strategies: (1) standard magnitude-based pruning, including Least Magnitude Pruning (LMP)  \citep{lwm} and its adversarially trained variant, AdvTrain+LMP; (2) pixel-level robust pruning, specifically HYDRA \citep{hydra} and {Flying Bird+} \citep{fbp}; and (3) sharpness-aware optimization, notably S$^2$-SAM \cite{s2sam}. \cite{adasap} is not open-sourced and is not included in our evaluation. 

\textit{Structured Pruning.} 
To demonstrate the versatility of the \sys{} framework, we further compare it against representative structured pruning methods, specifically DepGraph \citep{depgraph} and {HESSO} \citep{hesso}.
 
\noindent\textbf{Evaluation Metrics.} 
We evaluate performance using two primary metrics: (1) Task Accuracy and (2) Probabilistically Certified Accuracy (PCA), which quantifies the model's robustness against semantic perturbations. Following the CC-Cert framework \cite{cccert}, we randomly sample $m=100$ test instances and set the error bound $\varepsilon = 10^{-3}$ to compute PCA as follows:
\[
    \mathrm{PCA}(\mathcal{S}, \varepsilon) = \frac{1}{m}\sum_{i=1}^m \mathbf{1} \left[ \mathbb{P}_{x_{T}\sim\mathbb{S}_T(x_i)} \left( \hat{y}(x_T) = y_i \right) \ge 1 - \varepsilon \right].
\]
where $\mathcal{S}=\{(x_i, y_i)\}_{i=1}^m$ denotes the evaluation subset with $m=|\mathcal{S}|$, $\mathbf{1}[\cdot]$ is the indicator function, and $\hat{y}(\cdot)$ represents the prediction of the DNN classifier. 

Intuitively, a PCA of 80\% with $\varepsilon = 10^{-3}$ indicates that for 80\% of the test samples, the probability of the prediction flipping under semantic transformations is bounded by $10^{-3}$. Detailed descriptions of the certification framework and implementation are provided in \cref{appendix:cccert} and \cref{appendix:implementation}, respectively.

%% file: section/3-eval-res.tex
\begin{table*}[h]
\centering
\caption{Performance comparison of ResNet-18 across diverse datasets and augmentation scales.}
\label{tab:main_results}
\resizebox{0.9\textwidth}{!}{%
\setlength{\tabcolsep}{2.5pt} 
\renewcommand{\arraystretch}{1.05} 
\begin{tabular}{l|c|cccccc|cccccc}
\toprule
\multirow{3}{*}{\textbf{Method}} & \multirow{3}{*}{\textbf{Aug.}} & \multicolumn{6}{c|}{\textbf{Pruning Ratio = 50\%}} & \multicolumn{6}{c}{\textbf{Pruning Ratio = 70\%}} \\
\cmidrule(lr){3-8} \cmidrule(lr){9-14}
 & & \multicolumn{2}{c}{CelebA-HQ} & \multicolumn{2}{c}{Flowers-102} & \multicolumn{2}{c|}{CIFAR-10-C} & \multicolumn{2}{c}{CelebA-HQ} & \multicolumn{2}{c}{Flowers-102} & \multicolumn{2}{c}{CIFAR-10-C} \\
 & & Acc (\%)& PCA (\%)& Acc (\%)& PCA (\%)& Acc (\%)& PCA (\%)& Acc (\%)& PCA (\%)& Acc (\%)& PCA (\%)& Acc (\%)& PCA (\%)\\
\midrule
\midrule
\multirow{3}{*}{Vanilla Train} 
 & None & {83.62} & 50.0 & {97.80} & 58.0 & {93.04} & 26.0 & 83.62 & 50.0 & 97.80 & 58.0 & 93.04 & 26.0 \\
 & L1 & 82.96 & 66.0 & {97.56} & 84.0 & 91.84 & 80.0 & 82.96 & 66.0 & 97.56 & 84.0 & 91.84 & 80.0 \\
 & L2 & {83.79} & 71.0 & 96.70 & 81.0 & {93.32} & {86.0} & 83.79 & 71.0 & 96.70 & 81.0 & 93.32 & {86.0} \\
\cmidrule(lr){1-14} 
\multirow{3}{*}{AdvTrain} 
 & None & 72.26 & 2.0 & 90.33 & 28.0 & 73.09 & 0.0 & 72.26 & 2.0 & 90.33 & 28.0 & 73.09 & 0.0 \\
 & L1 & 72.59 & 12.0 & 88.14 & 32.0 & 73.37 & 2.0 & 72.59 & 12.0 & 88.14 & 32.0 & 73.37 & 2.0 \\
 & L2 & 72.02 & 14.0 & 86.80 & 26.0 & 75.05 & 1.0 & 72.02 & 14.0 & 86.80 & 26.0 & 75.05 & 1.0 \\
\midrule
\midrule
\multirow{3}{*}{LMP} 
 & None & \textbf{83.87} & 46.0 & 88.51 & 71.0 & 87.53 & 9.0 & 62.06 & 20.0 & 40.22 & 11.0 & 86.65 & 6.0 \\
 & L1 & \underline{82.47} & 54.0 & 89.24 & 71.0 & 86.47 & 51.0 & 63.70 & 39.0 & 70.90 & 46.0 & 86.02 & 48.0 \\
 & L2 & 81.98 & 66.0 & 87.90 & 55.0 & 86.63 & 56.0 & 67.90 & 46.0 & 69.93 & 36.0 & 84.64 & 46.0 \\
\cmidrule(lr){1-14}
\multirow{3}{*}{AdvTrain+LMP} 
 & None & 73.09 & 5.0 & 88.51 & 25.0 & 75.13 & 0.0 & 71.69 & 4.0 & 89.36 & 28.0 & 72.02 & 0.0 \\
 & L1 & 71.19 & 10.0 & 89.24 & 27.0 & 74.17 & 12.0 & 67.65 & 12.0 & 86.92 & 25.0 & 70.90 & 2.0 \\
 & L2 & 69.22 & 39.0 & 85.21 & 23.0 & 72.23 & 3.0 & 70.95 & 22.0 & 85.45 & 26.0 & 74.06 & 3.0 \\
\cmidrule(lr){1-14}
\multirow{3}{*}{HYDRA} 
 & None & 53.33 & 28.0 & 79.46 & 51.0 & 86.02 & 35.0 & 51.19 & 28.0 & 79.10 & 56.0 & 85.47 & 29.0 \\
 & L1 & 51.44 & 26.0 & 77.87 & 63.0 & 85.99 & 53.0 & 48.97 & 34.0 & 76.16 & 64.0 & 84.41 & 50.0 \\
 & L2 & 51.19 & 48.0 & 77.87 & 64.0 & 86.43 & 64.0 & 53.99 & 40.0 & 78.61 & 68.0 & 84.52 & 49.0 \\
\cmidrule(lr){1-14}
\multirow{3}{*}{Flying Bird+} 
 & None & 47.98 & 6.0 & 79.82 & 38.0 & 65.67 & 27.0 & 49.79 & 9.0 & 79.09 & 27.0 & 63.18 & 28.0 \\
 & L1 & 49.05 & 4.0 & 77.27 & 32.0 & 69.31 & 21.0 & 49.87 & 7.0 & 77.26 & 32.0 & 68.40 & 10.0 \\
 & L2 & 45.84 & 5.0 & 74.08 & 37.0 & 70.06 & 30.0 & 45.26 & 6.0 & 72.49 & 34.0 & 71.77 & 34.0 \\
\cmidrule(lr){1-14}
\multirow{3}{*}{S$^2$-SAM} 
 & None & 75.96 & 45.0 & 87.65 & 59.0 & 80.63 & 4.0 & 62.47 & 33.0 & 80.92 & 53.0 & 78.21 & 4.0 \\
 & L1 & 78.27 & 58.0 & 85.33 & 72.0 & 79.88 & 33.0 & 65.43 & 41.0 & 81.05 & 65.0 & 78.26 & 26.0 \\
 & L2 & 75.06 & 55.0 & 87.41 & 76.0 & 79.42 & 21.0 & 62.55 & 51.0 & 79.71 & 57.0 & 77.09 & 28.0 \\
\midrule
\rowcolor{gray!10} 
& L1 & {82.30} & \textbf{74.0} & \textbf{96.45} & \underline{88.0} & \underline{88.16} & \underline{60.0} & \underline{76.13} & \underline{56.0} & \underline{94.62} & \underline{79.0} & \underline{89.42} & \underline{57.0} \\
\rowcolor{gray!10} 
\multirow{-2}{*}{\textbf{\sys{}}} 
& L2 & {79.09} & \textbf{74.0} & \underline{96.33} & \textbf{91.0} & \textbf{92.70} & \textbf{82.0} & \textbf{76.46} & \textbf{63.0} & \textbf{96.70} & \textbf{90.0} & \textbf{91.77} & \textbf{67.0} \\
\bottomrule
\end{tabular}%
}
\end{table*}

\section{Empirical Results}\label{sec:results}
In this section, we conduct extensive experiments to evaluate the effectiveness of \sys{} and address the following research questions (RQs):

\noindent \textbf{RQ1.}Why do existing robust pruning approaches fail to keep robustness against semantic perturbations?

\noindent \textbf{RQ2.} How effective is \sys{} in improving PCA while maintaining task accuracy across different architectures and datasets?

\noindent \textbf{RQ3.} Does \sys{} generalize effectively to both structured and unstructured pruning schemes? 

\noindent \textbf{RQ4.} What are the  contributions of each component in the \sys{} composite objective to the overall performance? 

\subsection{Overall Performance Analysis (RQ1\&2)}
\cref{tab:main_results} details the performance on CelebA-HQ, Flowers-102, and CIFAR-10-C across various pruning ratios and augmentation scales. \sys{} consistently outperforms all baselines in the PCA metric. Addressing RQ1, we attribute the failure of existing methods to two factors.

First, magnitude-based methods (e.g., LMP, AdvTrain+LMP) assume that weight magnitude correlates with semantic importance. However, this heuristic ignores subnetwork geometric stability and leaves the semantic discrepancy $Z(x;T)$ unconstrained. Consequently, these models exhibit unstable PCA under increased augmentation. For instance, LMP on Flowers-102 (pr$=50\%$) suffers a 16\% PCA drop from L1 to L2.

Second, pixel-level robust pruning (e.g., AdvTrain+LMP, HYDRA, and Flying Bird+) aligns with off-manifold $\ell_{p}$ noise rather than the natural data manifold where semantic mutations occur. This misalignment results in poor robustness transfer, with PCA often falling below 10\%, highlighting the orthogonality between pixel-level and semantic robustness. While S$^{2}$-SAM improves stability by targeting flat minima, it still lags behind \sys{} due to the lack of explicit constraints on mask-induced variance.

For RQ2, \sys{} consistently achieves superior PCA across all experimental settings while maintaining task accuracy competitive with Vanilla Train. This success is rooted in our strategy of decomposing semantic discrepancy into compression-induced variance and intrinsic semantic sensitivity, which enables a more targeted optimization of model stability. By explicitly decoupling these factors, \sys{} can individually suppress the instability arising from mask optimization and the model's inherent sensitivity to transformations. Such efficacy is most pronounced in challenging high-compression regimes. For instance, on Flowers-102 (pr$=70\%$, L2 augmentation), \sys{} achieves 90\% PCA, surpassing the nearest competitor HYDRA by a substantial 22\% margin.

\begin{figure}[h]
    \centering
    \includegraphics[width=\linewidth]{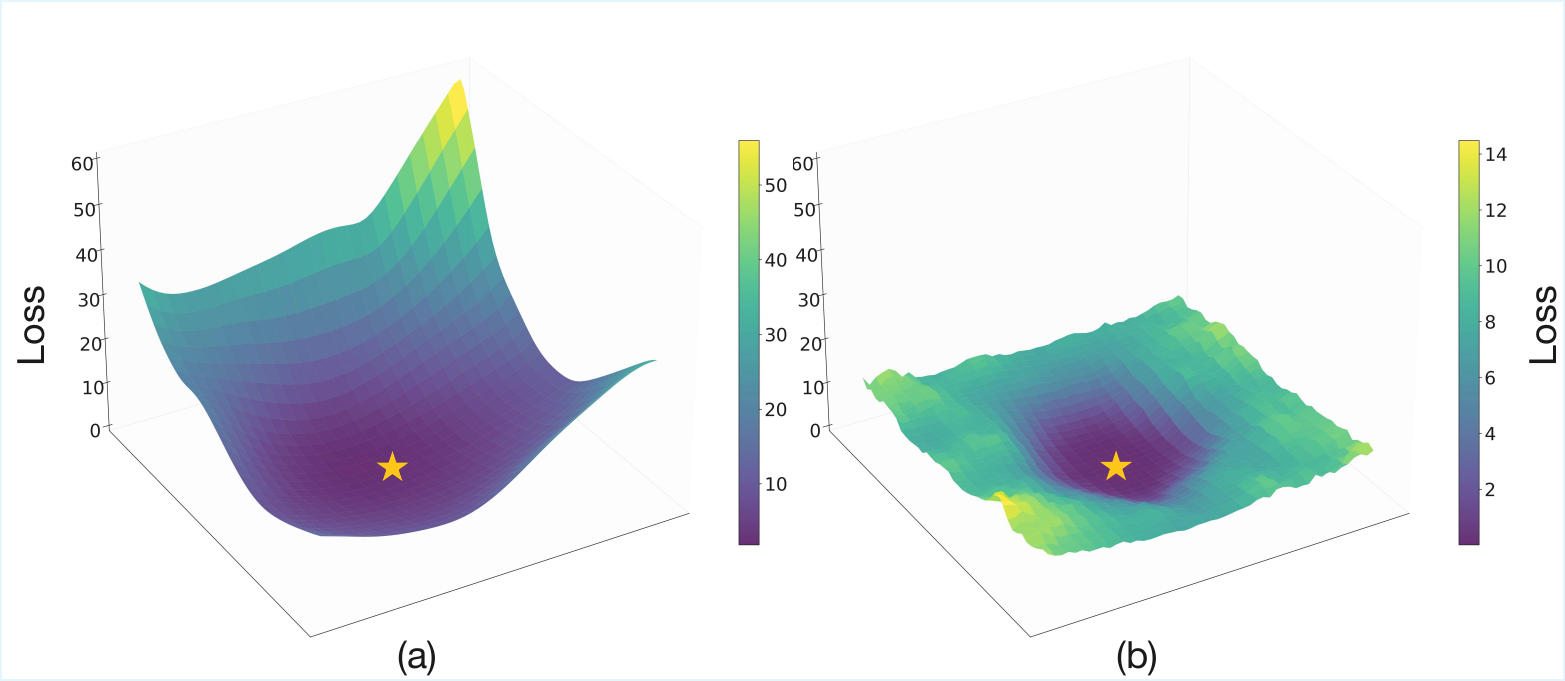}
    \caption{Loss landscape visualization for models trained with (a) adversarial loss and (b) \sys{} loss. The orange star indicates the location of the initial model. \vspace{-0.1in}}
    \label{fig:losslandscape}
\end{figure}

To better clarify the underlying mechanism, we visualize the loss landscapes in \cref{fig:losslandscape}. The model trained with conventional adversarial loss exhibits a sharp valley, where minor perturbations trigger significant loss increases, undermining robustness. In contrast, \sys{} yields a markedly flatter and smoother landscape. This geometric property aligns with established findings that link flat minima to enhanced robustness \citep{sharpness1,sharpness2,sharpness3,adasap,s2sam}.

\cref{tab:googlemobile} further extends this comparison to diverse architectures. Despite the structural differences among ResNet-18, GoogLeNet, and MobileNet-V2, \sys{} consistently delivers the highest certified robustness. The most striking improvement occurs in GoogLeNet, where \sys{} achieves 49.0\% PCA, outperforming the LMP baseline (7.0\%) by 42\%. Notably, \sys{} even surpasses vanilla unpruned models by 8\% to 27\%. This indicates that \sys{} effectively filters out non-robust redundant parameters that would otherwise degrade semantic stability.

\begin{table}[h]
\centering
\caption{Robustness generalization across architectures on CelebA-HQ (pr$=50\%$).}
\label{tab:googlemobile}
\resizebox{0.95\columnwidth}{!}{%
\setlength{\tabcolsep}{4pt}
\begin{tabular}{l|ccc|ccc}
\toprule
\multirow{2}{*}{\textbf{Model}} & \multicolumn{3}{c|}{{Acc (\%)}} & \multicolumn{3}{c}{{PCA (\%)}} \\
\cmidrule(lr){2-4} \cmidrule(lr){5-7}
 & Vanilla & LMP & \textbf{\sys{}} & Vanilla & LMP & \textbf{\sys{}} \\
\midrule
MobileNet-V2 & \textbf{71.85} & 33.91 & \underline{67.98} & \underline{33.0} & 6.0 & \textbf{44.0} \\
GoogLeNet    & \underline{67.16} & 44.12 & \textbf{70.86} & \underline{22.0} & 7.0 & \textbf{49.0} \\
ResNet-18    & \textbf{82.96} & \underline{82.47} & 79.09 & \underline{66.0} & 54.0 & \textbf{74.0} \\
\bottomrule
\end{tabular}}
\end{table} 

\subsection{Generalization across Pruning Schemes (RQ3)}
Beyond unstructured pruning, \cref{tab:structured_results} demonstrates that \sys{} generalizes effectively to structured pruning schemes.  Compared with DepGraph and HESSO, \sys{} consistently achieves higher PCA across all datasets and augmentation levels, confirming that the design is agnostic to the granularity of the prunable units and instead fundamentally tied to loss landscape geometry and semantic consistency. Specifically, on CelebA-HQ with L2 augmentation, \sys{} reaches 66\% PCA, whereas HESSO and DepGraph only manage 36\% and 60\%, respectively, validating \sys{}'s robustness in structured pruning scenarios.

\subsection{Ablation Study \& Hyperparameter Analysis (RQ4)}
\textbf{Component-wise Ablation Study.} 
\cref{tab:ablation} details the contribution of each component within the \sys{} objective. The removal of any single loss term causes a distinct PCA degradation, confirming that our components are complementary. Specifically, $\mathcal{L}_\text{stab}$ is essential for suppressing compression-induced variance, while $\mathcal{L}_{ratio}$ enforces semantic invariance relative to classification margins. Furthermore, $\mathcal{L}_{consis}$ is critical for bridging the soft-hard mask gap, ensuring training-phase robustness transfers to the deployed binary model. Notably, replacing KL divergence with MSE consistency further diminishes performance, indicating that aligning full predictive distributions is vital for preserving semantic robustness.

\begin{table}[htb]
\centering
\caption{Extension to Structured Pruning ($\text{pr}=50\%$). Comparison of accuracy and PCA across different augmentation levels.}
\label{tab:structured_results}
\resizebox{\linewidth}{!}{%
\setlength{\tabcolsep}{4pt} 
\renewcommand{\arraystretch}{1.1} 
\begin{tabular}{l|c|cc|cc|cc}
\toprule
\multirow{3}{*}{\textbf{Method}} & \multirow{3}{*}{\textbf{Aug.}} & \multicolumn{2}{c|}{\textbf{CelebA-HQ}} & \multicolumn{2}{c|}{\textbf{Flowers-102}} & \multicolumn{2}{c}{\textbf{CIFAR-10-C}} \\
\cmidrule(lr){3-4} \cmidrule(lr){5-6} \cmidrule(lr){7-8}
 & & Acc (\%) & PCA (\%) & Acc (\%) & PCA (\%) & Acc (\%) & PCA (\%) \\
\midrule
\multirow{3}{*}{DepGraph} 
 & None & 80.99 & 27.0 & 94.74 & 56.0 & \underline{92.10} & 20.0 \\
 & L1 & 79.51 & 50.0 & 94.87 & 67.0 & 90.98 & \underline{73.0} \\
 & L2 & 80.08 & 60.0 & 95.35 & 77.0 & \textbf{92.48} & 70.0 \\
\cmidrule(lr){1-8}
\multirow{3}{*}{HESSO} 
 & None & 56.54 & 30.0 & 81.78 & 57.0 & 85.26 & 17.0 \\
 & L1 & 58.18 & 43.0 & 80.92 & 73.0 & 84.54 & 61.0 \\
 & L2 & 55.39 & 36.0 & 80.80 & 68.0 & 84.77 & 67.0 \\
\midrule
\rowcolor{gray!10} 
& L1 & \textbf{85.72} & \underline{64.0} & \textbf{95.97} & \underline{80.0} & {90.21} & \textbf{76.0} \\
\rowcolor{gray!10} 
\multirow{-2}{*}{\textbf{\sys{}}} 
& L2 & \underline{83.21} & \textbf{66.0} & \underline{95.60} & \textbf{83.0} & {90.04} & \underline{73.0} \\
\bottomrule
\end{tabular}}
\end{table}

\begin{table}[htb]
\centering
\caption{Component-wise ablation study on CelebA-HQ. Performance degradation ($\downarrow$) is reported relative to the full \sys{} framework. \vspace{-0.1in}}
\label{tab:ablation}
\resizebox{0.95\columnwidth}{!}{%
\renewcommand{\arraystretch}{1.2}
\setlength{\tabcolsep}{6pt}
\begin{tabular}{l|c|cc}
\toprule
\textbf{Method / Variant} & {Acc (\%)} & {PCA (\%)} & {Drop} \\
\midrule
\rowcolor{gray!10} \textbf{\sys{}} & 82.72 & \textbf{74.0} & -- \\
\midrule
w/o Stability Loss ($\mathcal{L}_{\text{stab}}$) & 82.72 & 68.0 & \textcolor{black}{$\downarrow$ 6.0\%} \\
w/o Ratio Loss ($\mathcal{L}_{\text{ratio}}$) & 81.89 & 66.0 & \textcolor{black}{$\downarrow$ 8.0\%} \\
w/o Consistency Loss ($\mathcal{L}_{\text{consis}}$) & 83.37 & 67.0 & \textcolor{black}{$\downarrow$ 7.0\%} \\
w/ MSE Consistency & 82.14 & 65.0 & \textcolor{black}{$\downarrow$ 9.0\%} \\
w/o Sparsity Reg. ($\mathcal{L}_{1}$) & \textbf{83.05} & 71.0 & \textcolor{black}{$\downarrow$ 3.0\%} \\
\bottomrule
\end{tabular}}
\end{table}

\begin{table}[htb]
\centering
\caption{Impact of the initialization percentile $\tau$ on model performance. Results are evaluated on ResNet-18 using the CelebA-HQ dataset at pr$=50\%$. }
\label{tab:initmethod}
\resizebox{\columnwidth}{!}{%
\setlength{\tabcolsep}{4pt}
\renewcommand{\arraystretch}{1.2}
\begin{tabular}{l|c|cccc|c}
\toprule
\multirow{2}{*}{\textbf{Metric}} & \textbf{Random} & \multicolumn{4}{c|}{\textbf{Percentile-Scaled Initialization ($\tau\%$)}} & \textbf{Full-Scaled} \\
\cmidrule(lr){2-2} \cmidrule(lr){3-6} \cmidrule(lr){7-7}
 & (Baseline) & $\tau=40$ & $\tau=30$ & $\tau=20$ & $\tau=10$ & $\tau=0$ \\
\midrule
\textbf{Acc (\%)} & 78.11 & 80.82 & \textbf{82.72} & 82.55 & 83.05 & 0.49 \\
\textbf{PCA (\%)} & 59.0  & 63.0  & \textbf{74.0}  & 66.0  & 71.0  & 0.0 \\
\bottomrule
\end{tabular}%
}
\end{table}

\begin{figure}[htb]
    \centering
    \includegraphics[width=\linewidth]{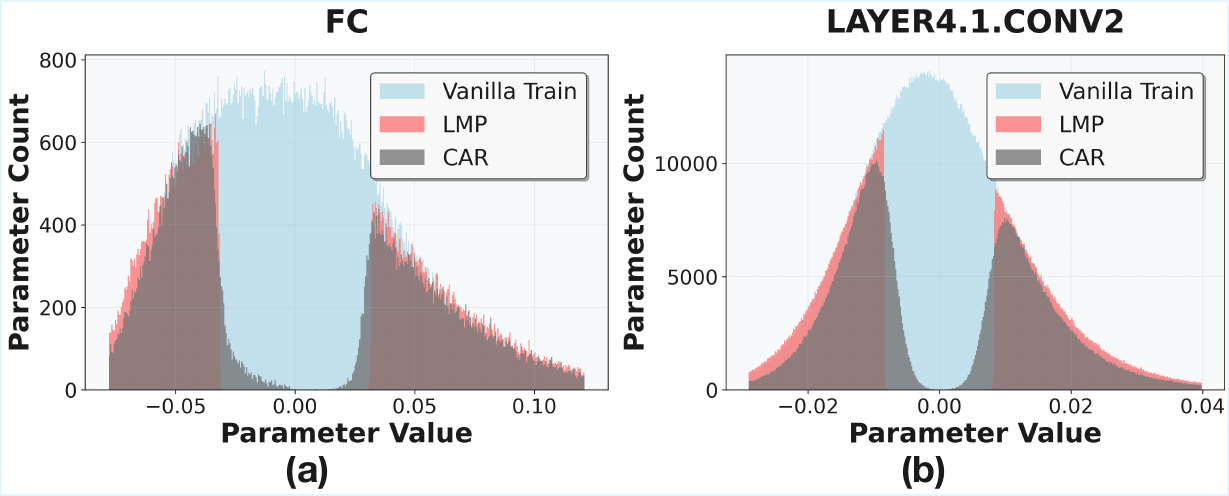}
    \caption{The parameter distributions for fully connected layer and \texttt{layer4.conv2}. }
    \label{fig:paradistribution}
\end{figure}

\begin{table}[htb]
\centering
\caption{Sensitivity analysis of training hyperparameters on ResNet-18 (CelebA-HQ, $\text{pr}=50\%$). The default settings are $\mu=0.5$ and $\eta=1.0$.}
\label{tab:hyperparams}
\resizebox{\columnwidth}{!}{%
\renewcommand{\arraystretch}{1.25}
\setlength{\tabcolsep}{8pt}
\begin{tabular}{l|ccc|ccc} 
\toprule
\multirow{2}{*}{\textbf{Metric}} & \multicolumn{3}{c|}{\textbf{Noise Magnitude ($\mu$)}} & \multicolumn{3}{c}{\textbf{Safety Margin ($\eta$)}} \\
\cmidrule(lr){2-4} \cmidrule(lr){5-7} 
 & 0.1 & 0.5  & 0.8 &  0.95 & 0.98 & 1.0  \\
\midrule
{Acc (\%)} & 82.55 & 82.72 & 82.30 & 82.06 & 82.55 & 82.72 \\
{PCA (\%)} & 70.0  & \textbf{74.0}  & 66.0   & 67.0  & 68.0  & \textbf{74.0} \\
\bottomrule 

\end{tabular}}
\end{table} 

\textbf{Hyperparameter Sensitivity Analysis.} Initialization of $C$ significantly influences optimization. We evaluate three strategies: (1) Random: $C$ values sampled from $\mathcal{U}(0,1)$; (2) Full-scaled: $C$ assigned proportional to weight magnitudes (equivalent to $\tau=0$); and (3) Percentile-scaled (Ours): $C$ initialized via \cref{eq:softmask} with $\tau \in \{10,20,30,40\}$. As shown in \cref{tab:initmethod}, full-scaled initialization fails to yield feasible results, while percentile-scaled with $\tau=30$ achieves the optimal trade-off, suggesting that extreme initialization constraints hinder the search for robust subnetworks.

\cref{fig:paradistribution} illustrates the weight distributions for both fully connected and convolutional layers. Notably, LMP removes nearly all parameters near zero, whereas \sys{} retains a moderate proportion of small-magnitude weights. Correlating this with \cref{tab:main_results}, preserving these parameters appears essential for balancing accuracy and certified robustness.

Beyond the initialization percentile $\tau$, we examine two pivotal hyperparameters: (1) $\mu$, controlling the mask noise intensity; and (2) $\eta$, the safety margin in $\mathcal{L}_{ratio}$. \cref{tab:hyperparams} details the sensitivity of task accuracy and PCA to these parameters.
A moderate noise level ($\mu=0.5$) achieves the peak PCA of 74\%, indicating that optimal stochasticity regularizes the subnetwork to enhance robustness. Conversely, excessive noise (e.g., $\mu=0.8$) destabilizes optimization. Regarding the safety margin, $\eta=1.0$ yields the best performance (74\% PCA). Lower values impose overly aggressive constraints, which impair the subnetwork's representational capacity and final stability.

%% file: section/5-conclude.tex

\section{Conclusion}\label{sec:conclusion}

This work presents \sys{}, a unified training framework for obtaining compact and certified robust DNNs against semantic perturbations. 
By optimizing pruning masks through a composite objective that incorporates structural flatness and semantic consistency, \sys{} effectively bridges the gap between discrete model pruning and continuous robust optimization. 
Extensive evaluations across multiple architectures and datasets demonstrate that \sys{} consistently achieves superior certified robustness, with improvements up to 42\% over existing baselines, while maintaining task accuracy. 
Future work will explore \sys{}’s interplay with multi-attribute semantic variations using more advanced generative models, and further reduce its computational overhead for optimizing certified robustness. 




%% file: section/6-appendix.tex
\appendix
\onecolumn
\section{Appendix A}
\label{appendixa}

\subsection{Proof for $Z(x;T)$}\label{appendix:proof_Z}
\begin{proof}
To derive a manageable upper bound for the prediction discrepancy $Z(x;T)$ under a specific mask instance $C$, we utilize the triangle inequality and the relationship between vector norms.

Let $\mathbf{p}_{C}(x)$ be the probability vector generated by the model with mask $C$, and let $\bar{\mathbf{p}}(x) = \mathbb{E}_{\xi}[\mathbf{p}_{C_\xi}(x)]$ denote the ensemble mean over all possible stochastic mask instances. We decompose the total discrepancy by adding and subtracting the ensemble means for both the original input $x$ and the transformed input $x_T$:
\begin{align}
Z(x;T)
&=\big\|\mathbf{p}_{C}(x)-\mathbf{p}_{C}(x_T)\big\|_\infty \nonumber\\
&=\big\|\mathbf{p}_{C}(x)-\bar{\mathbf{p}}(x) + \bar{\mathbf{p}}(x) - \bar{\mathbf{p}}(x_T) + \bar{\mathbf{p}}(x_T) -\mathbf{p}_{C}(x_T)\big\|_\infty \nonumber\\
&\le \big\|\mathbf{p}_{C}(x)-\bar{\mathbf{p}}(x)\big\|_\infty
+\big\|\bar{\mathbf{p}}(x)-\bar{\mathbf{p}}(x_T)\big\|_\infty +\big\|\bar{\mathbf{p}}(x_T)-\mathbf{p}_{C}(x_T)\big\|_\infty \nonumber \\
& \le \underbrace{\sqrt{\big\| \mathbf{p}_{C}(x) - \bar{\mathbf{p}}(x) \big\|_2^2}}_{\mathrm{(A)}}
+ \underbrace{\big\| \bar{\mathbf{p}}(x) - \bar{\mathbf{p}}(x_T) \big\|_\infty}_{\mathrm{(B)}}  + \underbrace{\sqrt{\big\| \bar{\mathbf{p}}(x_T) - \mathbf{p}_{C}(x_T) \big\|_2^2}}_{\mathrm{(C)}}\nonumber
\end{align}

\end{proof}
\subsection{Proof for Label Invariance}\label{proof:lableinvariance}
Let $y$ be the true label of input $x$. We assume the classifier correctly predicts the input $x$, i.e., $c=\arg \max \mathbf{p}(x) = y$. We denote $c=\arg\max \ \mathbf{p}$ and $\tilde{c}=\arg\max \ \mathbf{p}_T$. Given the assumption $c=y$, if $\|\mathbf{p}(x)-\mathbf{p}(x_T)\|_\infty < d=\frac{p_1-p_2}{2}$ holds, then $\tilde{c}=c=y$.
\begin{proof}
Assume, for contradiction, that $\tilde{c} \neq c$.  
In this case one has 
\[
p_{T\tilde{c}} > p_{Tc}, \qquad p_c > p_{\tilde{c}} .
\]

Moreover, the condition $\|\mathbf{p}-\mathbf{p}_T\|_\infty < d$ implies that 
\[
|p_k - p_{Tk}| < d \quad \text{for all } k \in \{1,\dots,K\}.
\]
In particular, for indices $c$ and $\tilde{c}$ it follows that
\[
p_{Tc} > p_c - d, \qquad p_{T\tilde{c}} < p_{\tilde{c}} + d .
\]

Consequently,
\[
p_{Tc} - p_{T\tilde{c}} \;>\; (p_c - d) - (p_{\tilde{c}} + d) 
= p_c - p_{\tilde{c}} - 2d .
\]

On the other hand, the assumption $p_{T\tilde{c}} > p_{Tc}$ requires $p_{Tc} - p_{T\tilde{c}} < 0$.  
This contradicts the fact that the right-hand side above is non-negative whenever $p_c - p_{\tilde{c}} \geq 2d$.  
Hence, the assumption $\tilde{c} \neq c$ is invalid, and we conclude that $\tilde{c} = c$.
\end{proof}

\subsection{Variance Identity}\label{variance identity}
\begin{proof}
Let $C_m$ and $C_n$ be two independent stochastic mask instances derived from $C$ via the noise injection process defined in \cref{eq:mask}, i.e., $C_\xi = \text{clip}(C + \xi, 0, 1)$ with $\xi \sim \mathcal{U}(-\mu, \mu)$. For a fixed input $x$, let $\mathbf{p}_{C_\xi} = f(x; \mathcal{T}(C_\xi) \odot \theta)$, and $\bar{\mathbf p}=\mathbb E_\xi[\mathbf p_{C_\xi}]$ denote the ensemble mean. Then:
\begin{align*}
\mathbb E_{C_m,C_n}\big[\|\mathbf p_{C_m}-\mathbf p_{C_n}\|_2^2\big]
&= \mathbb E_{C_m,C_n}\big[\langle \mathbf p_{C_m}-\mathbf p_{C_n}, \mathbf p_{C_m}-\mathbf p_{C_n} \rangle\big] \\
&= \mathbb E_{C_m}\|\mathbf p_{C_m}\|_2^2 + \mathbb E_{C_n}\|\mathbf p_{C_n}\|_2^2
   - 2\,\mathbb E_{C_m,C_n}[\langle \mathbf p_{C_m}, \mathbf p_{C_n}\rangle] \\
&\quad \text{(Since $C_m$ and $C_n$ are independent, $\mathbb E[XY] = \mathbb E[X]\mathbb E[Y]$)} \\
&= 2\,\mathbb E_\xi\|\mathbf p_{C_\xi}\|_2^2
   - 2\,\langle \mathbb E_{C_m}[\mathbf p_{C_m}],\ \mathbb E_{C_n}[\mathbf p_{C_n}]\big\rangle \\
&= 2\,\mathbb E_\xi\|\mathbf p_{C_\xi}\|_2^2 - 2\,\langle \bar{\mathbf p}, \bar{\mathbf p} \rangle \\
&= 2\,\mathbb E_\xi\|\mathbf p_{C_\xi}\|_2^2 - 2\,\|\bar{\mathbf p}\|_2^2 \\
&= 2\,\big(\mathbb E_\xi\|\mathbf p_{C_\xi}\|_2^2 - 2\|\bar{\mathbf p}\|_2^2 + \|\bar{\mathbf p}\|_2^2 \big) \\
&= 2\,\big(\mathbb E_\xi\|\mathbf p_{C_\xi}\|_2^2 - 2\langle \mathbb E_\xi[\mathbf p_{C_\xi}], \bar{\mathbf p} \rangle + \|\bar{\mathbf p}\|_2^2 \big) \\
&= 2\,\mathbb E_\xi\big[\|\mathbf p_{C_\xi}\|_2^2 - 2\langle \mathbf p_{C_\xi}, \bar{\mathbf p} \rangle + \|\bar{\mathbf p}\|_2^2\big] \\
&= 2\,\mathbb E_\xi\big[\|\mathbf p_{C_\xi}-\bar{\mathbf p}\|_2^2\big].
\end{align*}
\end{proof}

\subsection{Detailed Description of Dataset and Augmentations}\label{appendix:data}
This section details the source and methods of how we create the training dataset.

\noindent\textbf{Dataset Introduction.} 
(1) {CelebA-HQ}~\citep{dataset}: A facial identity dataset comprising 307 identities (4,263 training, 1,215 testing images). We employ DIFF-AE~\citep{diffae} to introduce semantic variations, specifically transforming faces to have ``smiling" expressions.
(2) {Oxford Flowers-102}~\citep{flowers}: A dataset of 102 flower categories (6,552 training, 818 testing images). We utilize DiffusionCLIP~\citep{Diffusionclip} to generate mutated images based on the prompt ``A flower in focus, with the sun rising behind it, casting a warm golden glow."
(3) {CIFAR-10}~\citep{cifar10}: To evaluate the generalization of \sys{} beyond generative models, we adopt the ``fog" corruption (Severity Level 2) from the CIFAR-10-C benchmark~\citep{cifar10-c}.

To evaluate the scalability of \sys{} against off-manifold perturbations, we incorporate the ``fog" corruption from the CIFAR-10-C benchmark \cite{cifar10-c}. Unlike the diffusion-based semantic mutations used for CelebA-HQ and Flowers-102, the transformation space for CIFAR-10-C is constructed via pixel-level linear interpolation. Specifically, we define the transformation function $T$ as follows:
\begin{equation}
T(x;\delta) = \text{Clip}\Big(\text{LinearInterp}\big(x, Fog(x, L_2), \delta\big), 0, 1\Big), \quad \delta \in [0, 1]
\end{equation}
where $Fog(\cdot)$ denotes the foggy augmentation algorithm at severity level 2. The parameter $\delta$ controls the interpolation intensity, transitioning from the original image ($\delta=0$) to the fully corrupted version ($\delta=1$). During the certification process, we sample $\delta_i \sim \mathcal{U}(0, 1)$ to obtain the set of transformed inputs for estimating the upper bound of the prediction discrepancy $Z$. 

\noindent\textbf{Augmentation Settings.} 
For the diffusion-based datasets, we set the perturbation magnitude $\|\delta\|=1$ to produce the largest semantic shift. To analyze the impact of data scale, we define two augmentation levels: we augment CelebA-HQ and Flowers-102 with {1,000} and {4,000} mutated samples, respectively. For CIFAR-10, we scale these levels to {10,000} and {20,000} fog-corrupted samples.

\subsection{Detailed Description of Baselines}\label{appendix:baseline}
This section provides the technical definitions and strengthens the contrast between these methods and our proposed \sys{} framework.

\noindent\textbf{Robustness and Unstructured Pruning Baselines.}
We compare \sys{} against standard and robust training strategies, as well as methods that integrate pruning with adversarial defense:
\begin{itemize}
    \item \textbf{Vanilla Train}: The standard baseline where the model is trained using only cross-entropy loss without any pruning or robust training.
    \item \textbf{AdvTrain}: A standard defense method that incorporates adversarial training~\citep{pgd} to optimize robustness against pixel-level $L_p$-norm perturbations, ignoring semantic-level variations.
    \item \textbf{LMP (Least Magnitude Pruning)}~\citep{lwm}: A classic magnitude-based pruning approach that removes weights with the smallest absolute values, disregarding the geometric properties of the loss landscape and the topology of the subnetwork.
    \item \textbf{AdvTrain+LMP}: A sequential combination that applies LMP to an adversarially trained model, representing a naive attempt to combine compactness and robustness.
    \item \textbf{HYDRA}~\citep{hydra}: A representative robust pruning framework that optimizes the pruning mask using the TRADES~\citep{trades} adversarial loss, yet remains limited to pixel-level robustness objectives.
    \item \textbf{Flying Bird+}~\citep{fbp}: A dynamic sparsity method that initializes a random sparse subnetwork and jointly optimizes weights and topology, focusing on efficient training rather than certified stability.
    \item \textbf{S$^2$-SAM}~\citep{s2sam}: A sharpness-aware minimization approach designed for sparse models. While it steers optimization toward a smooth loss landscape to enhance generalizability, it lacks explicit constraints for semantic robustness. In practice, S$^2$-SAM is used after magnitude-based one-shot pruning to finetune the sparse model.
\end{itemize}
For the adversarial attack used in experiments, we employ an $L_2$-norm PGD attack, setting the perturbation budget to $\epsilon=0.5$ for CelebA and CIFAR-10, $\epsilon=2.0$ for Flowers. 

\noindent\textbf{Structured Pruning Baselines.}
To evaluate the versatility of \sys{} in structured pruning settings, we compare with advanced structured pruning methods:
\begin{itemize}
    \item \textbf{DepGraph}~\citep{depgraph}: A method that constructs a dependency graph to group coupled parameters, ensuring structural consistency but without robustness guarantees.
    \item \textbf{HESSO}~\citep{hesso}: A most recent structured pruning technique from the "Only-Train-Once" (OTO)~\citep{otov1,otov2} family, which focuses purely on training efficiency and structural redundancy removal.
\end{itemize}

\subsection{Certification Framework: CC-Cert}\label{appendix:cccert}
According to \cref{eq:z} and \cref{eq:cpp}, the label invariance condition is $Z<d$. Hence, our goal is to bound $P(Z \ge d)$ from the above. Using Markov's inequality:
\begin{equation}
    \mathbb{P}(Z \geq t) \leq \frac{\mathbb{E}(Z)}{t},
\end{equation}
where $Z$ is a non-negative random variable and $t \in \mathbb{R}^+$. This provides an initial bound, but CC-Cert refines this using the Chernoff-Cramer inequality \citep{cc}:
\begin{equation}\label{cc-eq}
\mathbb{P}(Z \geq d)=\mathbb{P}(e^{Zt} \geq e^{dt}) \leq \frac{\mathbb{E}(e^{Zt})}{e^{dt}}.
\end{equation}
This gives a tighter upper bound for $\mathbb{P}(Z \ge d)$, which serves as an upper bound for $\varepsilon$ in \cref{1-epsilon}. The optimal value of $t$ is chosen to minimize this bound. However, the true expectation of $e^{Zt}$ is difficult to compute directly.

To address this, CC-Cert \citep{cccert} estimates $\mathbb{E}(e^{Zt})$ by sampling $n$ transformed inputs $\{x_{T_i}\}_{i=1}^n$, calculating $Z_i=\|\mathbf{p} - \mathbf{p}_{T_i}\|_\infty$ for each, and using:
\begin{equation}\label{one_bound_Y}
    Y=\frac{1}{n\cdot e^{dt}}\sum_{i=1}^{n} e^{Z_i t}.
\end{equation}
$Y$ is an estimate of the upper bound on $\varepsilon$. While $Y$ could overestimate or underestimate the true value, underestimation is more problematic as it could lead to false certification. To mitigate this risk, we sample $Y$ independently $l$ times, resulting in $\{Y_1, \dots, Y_l\}$. The probability of underestimation is then bounded using the following inequality (derived from \cite{cccert, lemma1}):

\begin{equation}\label{eq:paley}
    \mathbb{P}\left(\frac{\max\{Y_1,\dots,Y_l\}}{\alpha }< \frac{\mathbb{E}(e^{Xt})}{e^{dt}}\right) < \left(\frac{1}{1+\frac{n(1-\alpha)^2}{C_v^2}}\right)^l,
\end{equation}
where $X$ is a random variable in $[0,1]$, $\alpha$ is a hyperparameter, and $C_v=\frac{\text{Var}(e^{Xt})}{\mathbb{E}(e^{Xt})} \sim 1$ is a coefficient related to $e^{Xt}$.

To obtain a reliable estimate of the upper bound on $\varepsilon$, we repeat the following process $l$ times. For each repetition, we sample $n$ transformed images $\{x_{T_i}\}_{i=1}^n$ from the semantic transformation set $\mathbb{S}_T$ for a given input $x$, compute $Z_i=\|\mathbf{p} - \mathbf{p}_{T_i}\|_\infty$ for each, and calculate the corresponding estimate $Y_i$ using \cref{one_bound_Y}. The maximum of these $l$ independent estimates, $\max\{Y_1, \dots, Y_l\}$, is then used as a conservative upper bound for $\varepsilon$. This ensures that, for sufficiently large $n$ and $l$, the probability of underestimating the true $\varepsilon$ (and therefore falsely certifying a non-robust model) becomes arbitrarily small. This provides a high-confidence probabilistic robustness guarantee.

For the settings, we set $\alpha=0.9$, $n=100$, and $l=10$ in \cref{eq:paley}, and sweep the temperature vector $t$ logarithmically over $[10^{-4},10^{4}]$ with 500 intervals.

\subsection{Implementation Details}\label{appendix:implementation}
The training pipeline proceeds in three stages. 
First, the initial pretraining stage is conducted for 50 epochs with a learning rate of 0.01. 
Next, the compression-aware robustness optimization runs for 100 epochs with a learning rate of 0.0001, utilizing the hyperparameter settings $\lambda_\text{stab}=5$, $\lambda_\text{ratio}=1$, $\lambda_\text{consis}=1$, and $\lambda_\text{1}=0.0001$. 
Finally, the compressed model is fine-tuned for 50 epochs with a learning rate of 0.001. 
All experiments are implemented in PyTorch and executed on a single NVIDIA RTX A6000 GPU.

\newpage
\subsection{Illustration of Transformation Space}
\begin{figure}[h]
    \centering
    \includegraphics[width=1\linewidth]{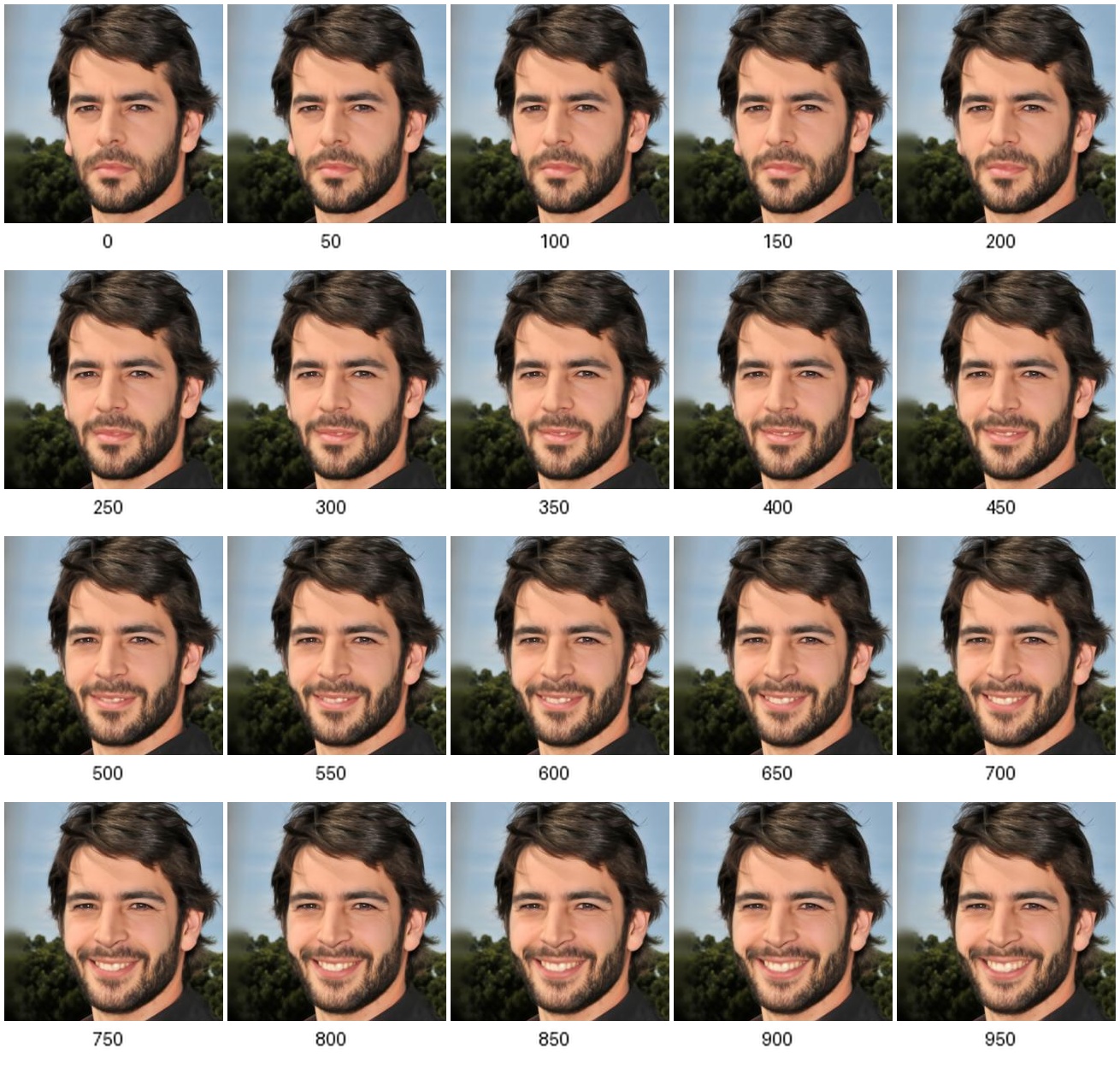}
    \caption{The demonstration of the transformation space of $\mathbb{S}_T(x)$, where $x$ is the original image, labeled 0.}
    \label{fig:ccert_grid}
\end{figure}